\documentclass{article} 
\usepackage{nips13submit_e,times}
\usepackage{hyperref}
\usepackage{natbib}
\usepackage{url}
\usepackage{graphicx}
\usepackage{caption}
\usepackage{subcaption}
\usepackage{tabulary}
\usepackage[utf8]{inputenc}

\title{Predicting State-Level Agricultural Sentiment with Tweets from Farming Communities}

\author{
\large{Jared Dunnmon, Swetava Ganguli, Darren Hau, Brooke Husic} \\
\large{\texttt{\{jdunnmon, swetava, dhau, bhusic\} @ cs.stanford.edu}} \\
\large{Stanford University}\\
}

\nipsfinalcopy 

\begin{document}

\maketitle

\begin{abstract}
The ability to obtain accurate food security metrics in developing areas where relevant data can be sparse is critically important for policy makers tasked with implementing food aid programs. As a result, a great deal of work has been dedicated to predicting important food security metrics such as annual crop yields using a variety of methods including simulation, remote sensing, weather models, and human expert input.  As a complement to existing techniques in crop yield prediction, this work develops neural network models for predicting the the sentiment of Twitter feeds from farming communities.  Specifically, we investigate the potential of both direct learning on a small dataset of agriculturally-relevant tweets and transfer learning from larger, well-labeled sentiment datasets from other domains (e.g.~politics) to accurately predict agricultural sentiment, which we hope would ultimately serve as a useful crop yield predictor.  We find that direct learning from small, relevant datasets outperforms transfer learning from large, fully-labeled datasets, that convolutional neural networks broadly outperform recurrent neural networks on Twitter sentiment classification, and that these models perform substantially less well on ternary sentiment problems characteristic of practical settings than on binary problems often found in the literature.
\end{abstract}

\section{Introduction}
Obtaining reliable data describing local food security metrics at a granularity that is informative to policy-makers requires expensive and logistically difficult surveys, particularly in the developing world. Predicting crop yield in developing countries, for instance, is a valuable tool for designing policies to address food security and poverty.  However, current methods such as in-person expert evaluation, weather modeling, and traditional econometric techniques do not scale well in geographically inaccessible or poorer areas.  Inspired by work done by Jean et al. \cite{jean2016} and You et al. \cite{you2017}, our group has previously worked to predict food security predictions applying Convolutional Neural Networks (CNNs) to satellite imagery, which demonstrated a promising ability to directly predict metrics such as stunting percentage \cite{cs2212016}. In addition, using convolutional neural networks and GANs on geospatial data in unsupervised or semi-supervised settings has also been of interest recently; especially in domains such as food security, cybersecurity, satellite tasking, etc. (\cite{ganguli2019predicting, ganguli2019geogan, perez2019semi}). We believe that the predictive power of our model could be improved by incorporating additional information from text-based sources that incorporate high-granularity knowledge from on-the-ground observation. Such sources could be social media websites like Twitter, Tumblr and Facebook, or newsfeeds from various outlets.  Regardless of source, the key challenge in utilizing such textual data is the fact that fully-labeled datasets related to sentiment about crop yield or agricultural performance are effectively nonexistent. 
While substantial amounts of labeled sentiment data exist in other domains such as restaurant reviews (Yelp), hotel reviews (TripAdvisor), movie reviews (IMDB), the ability of sentiment models to transfer accurately from these training distributions to a test distribution involving agriculturally-relevant data is unclear. 
Thus, in this work, we investigate two parallel routes for predicting agricultural sentiment from Twitter data.  The first of these involves using a small, domain-relevant dataset to predict on a held-out test set containing agriculturally-relevant tweets.  The underlying hypothesis here is that the ontology of Twitter is extremely domain-specific to the point that a small, but relevant training set would enable the creation of a useful domain-specific sentiment classification model.



In the spirit of transfer learning models that have been successfully applied in such fields as computer vision (\cite{saenko10}, \cite{Lim11}), our second hypothesis states that we can overcome the challenge of sparse datasets by leveraging rich, fully-labeled datasets curated for other purposes, such as sentiment prediction on political learnings, economic health, and movie reviews.  Sentiment analysis on these easily available datasets, although unrelated to the domain of interest, may encode features and learnings that are transferable to smaller target datasets.  


To test these hypotheses, we (1) located and obtained several rich, fully-labeled sentiment analysis datasets for unrelated domains from websites such as CrowdFlower; (2) extracted and labeled Twitter posts related to crops, location, and weather. Approximately 12 man-hours was spent by each team-member in the process of manually labeling our dataset; (3) trained several different NLP models (e.g.~GRU variants, LSTM variants, CNN) on fully-labeled and augmented datasets and performed thorough hyperparameter searches; and (4) evaluated model performance on a held-out test dataset directly applicable to crop yield prediction.


\section{Background and Related Work}
\textit{Crop Yield}---Traditional methods for predicting crop yield are described by Porter et al \cite{porter2014}.  Most established methods use models incorporating a subset of precipitation, temperature, CO$_2$ and O$_3$ (ozone), and soil parameters, which are not always readily available, particularly in developing areas of the world where food security data is most important.  To overcome this challenge, You et al.~leverage remote sensing imagery and state-of-the-art machine learning techniques to automatically extract features of interest \cite{you2017}.  They process a time series of multi-spectral images into a time series of histograms of pixel intensities.  These histograms can be stacked and fed into a CNN with convolution over the “bin” and “time” dimensions, or they can be fed into an LSTM.  You et al. show that these models significantly outperform widely used crop yield prediction models such as ridge regression, decision trees, and dense neural networks. In CS 221, our group built a similar pipeline for predicting food security metrics.  Instead of deriving histograms to be fed into CNNs and LSTMs, we directly applied a CNN to satellite images to predict stunting percentage metrics for different geographic clusters in Nigeria.  With a hyperparameter search performed only for learning rate, we achieved above 40\% accuracy (substantially above the baseline) on classifying images by food security metric levels.  Our belief that incorporating additional information from near real-time text sources could supplement this imagery-based approach has motivated the current work.

\textit{Sentiment Analysis}---Sentiment analysis is a popular topic in the natural language processing field, although it is typically applied to areas such as political sentiment and movie reviews.  Peter Nagi, for instance, \cite{nagi2017} uses LSTMs for sentiment analysis of tweets from the first GOP primary debate.  After removing the neutral examples such that he predicts only positive and negative classes, Nagi achieves a test accuracy of 84\%, which decomposes into 93\% accuracy for negative tweets and 57\% accuracy for positive tweets.  He hypothesizes that these unbalanced results occur because the GOP debate tweets featured a substantially greater number of negatively labeled examples than positively labeled examples.

Socher et al.~\cite{socher2013} use the Rotten Tomatoes dataset, originally compiled by Pang and Lee \cite{pang05}, in their Stanford Sentiment Treebank, in which sentences are parsed into a tree structure.  They pioneer the use of a Recursive Neural Tensor Network (RNTN) and evaluate its performance on sentiment classification on all nodes (words and phrases in sentence) and on the root node (entire sentence), for both fine-grained sentiments of five classes (very negative, negative, neutral, positive, and very positive) and coarse positive/negative classes.  The RNTN achieves improved accuracies across the board compared with existing models, and it works particularly well with shorter text, given the increased importance of negation and composition.  It is worth noting, however, that predicting fine-grained sentiments of entire sentences is much more difficult than any of the other tasks.

Yoon Kim \cite{kim2014} adapts CNNs for sentence classification on a variety of tasks (i.e. sentiment analysis, question type).  This work suggests that a single layer CNN is comparable to or performs marginally better than state-of-the-art models for each of the tasks considered. Given the success of the models described here, we propose using variants of GRU and LSTM (e.g.~single cell, bi-directional, multi-layer) models in addition to CNN-based architectures to assess the ability of both direct learning and transfer learning to predict agricultural sentiment that would help to accurately predict crop yield.

\textit{Transfer Learning}---Previous work by Bartusiak et al.~\cite{bartusiak2015} demonstrates that transfer learning can be effective, even for a simple SVM classifier.  We hypothesize that by consistently using GloVe word embeddings that capture the semantic similarities among words, we could enable effective cross-domain transfer learning on sentiment prediction tasks.

\section{Data}



Fundamentally, we are interested in assessing whether models achieve better performance predicting agricultural sentiment when those models are trained on (a) a smaller, domain-relevant dataset or (b) a richer dataset from one or more unrelated domains. In other words, we want to know if a sentiment model trained on non-agricultural inputs can generalize to our domain of interest. Descriptions of our small, domain-relevant dataset and our richer, unrelated datasets are provided below.

\textit{Target Dataset on Agricultural Sentiment}---To obtain the smaller target dataset, we designed a Twitter query to extract tweets of interest, which we defined as containing (i) crop of interest, e.g.~``wheat'', ``lettuce'', ``soybeans'' based on major U.S. agricultural products \cite{wikipedia}; (ii) state name, e.g.~``Kansas'', ``Nebraska''; and (iii) a word describing weather, e.g.~``rain'', ``hail''.  This query returned nearly 100,000 tweets from the year 2016. After filtering for relevance and deleting identical tweets, we retained approximately 4000 tweets.  We manually labeled this dataset with five sentiment classes: 0 - very negative, 1 - negative, 2 - neutral, 3 - positive, 4 - very positive.  After a final manual filter that eliminated duplicates and irrelevant tweets, our dataset consisted of about 2300 tweets suitable for analysis.  This exercise itself (i.e.,~obtaining and manually labeling a specialized dataset) demonstrates the potential value of our project, given the effort required to obtain just 2300 relevant examples.

\begin{figure}
    \centering
    \begin{subfigure}[b]{0.3\textwidth}
        \includegraphics[width=\textwidth]{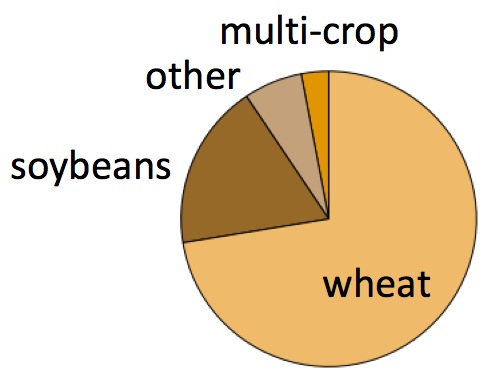}
        \caption{By crop}
        \label{fig:by_crop}
    \end{subfigure}
    \hspace{0.2cm}
    \begin{subfigure}[b]{0.3\textwidth}
        \includegraphics[width=\textwidth]{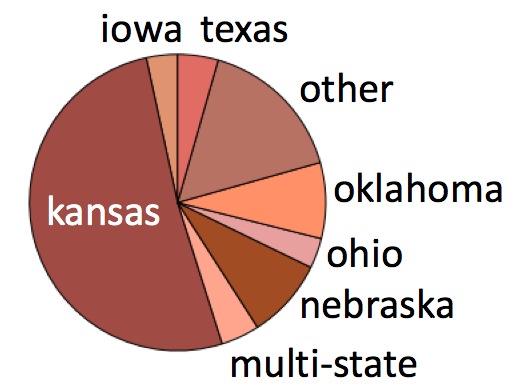}
        \caption{By state}
        \label{fig:by_state}
    \end{subfigure}
    \caption{Distribution of agriculture-related tweets by crop and by state}\label{fig:ag_dist}
\end{figure}

The breakdown of the tweets revealed that the most common location was Kansas, and the most common crop was wheat (see Fig.~\ref{fig:ag_dist}). Therefore, we chose our target dataset for final agricultural sentiment prediction to be all tweets about wheat in Kansas between July and December 2016. This set consisted of 266 tweets and was isolated from all model training. For training, we used all (suitable) tweets posted before July about any crop and any state, which amounted to 882 tweets. This design ensures that we are not contaminating our final prediction with data about other states and crops that occurred during the same time interval as our isolated Kansas wheat dataset.

In summary, this twitter query produced two datasets that we will refer to by the following abbreviations:

\begin{enumerate}
\item [KWT] Kansas Wheat Test: 266 tweets about wheat in Kansas from July 2016 -- December 2016
\item [AG] Agricultural sentiment: 882 tweets about crops in the U.S.~from March 2016 -- June 2016.
\end{enumerate}

\textit{Prelabeled Sentiment Datasets}---We used three distinct prelabeled sentiment datasets in order to train sentiment models on domains unrelated to agriculture. All three datasets contain only small sentences (52 words are fewer) and two of them are from Twitter. These fully-labeled datasets are as follows:

\begin{enumerate}
\item [GD] Tweets about the first 2016 GOP primary debate (13871 examples, 3 sentiment classes).\cite{crowdflower}
\item [SDC] Tweets about self-driving cars (6943 examples, 5 sentiment classes)\cite{crowdflower}
\item [RT] Movie reviews from Rotten Tomatoes (11855 examples, 0-1 continuous sentiment classification) \cite{SST}
\end{enumerate}

To improve cross-domain performance, we mapped each labeled dataset (KWT, AG, GD, SDC, and RT) to three sentiments (negative, neutral, and positive). Then, we combined these datasets into larger, fully-labeled datasets as follows:

\begin{enumerate}
\item [TO] Twitter Only: GD + SDC datasets
\item [FC] Full Combo: GD + SDC + RT datasets
\item [TOA] Twitter Only with Agriculture: GD + SDC + AG datasets
\item [FCA] Full Combo with Agriculture: GD + SDC + RT + AG datsets
\end{enumerate}

All datasets were reasonably balanced with respect to distribution of negative, neutral, and positive sentiment.

\begin{center}
\begin{tabular}{ |c|c|c|c| }
    \hline
    Dataset/Sentiment & Negative & Neutral & Positive \\
    \hline
    Agriculture & 38.8\% & 32\% & 29.3\% \\
    Twitter Only & 44.6\% & 35.5\% & 19.9\% \\
    Full Combo & 40.4\% & 33.9\% & 25.7\% \\
    \hline
\end{tabular}
\end{center}

\section{Approach}

\begin{figure}
	\centering
	\includegraphics[width=0.5\textwidth]{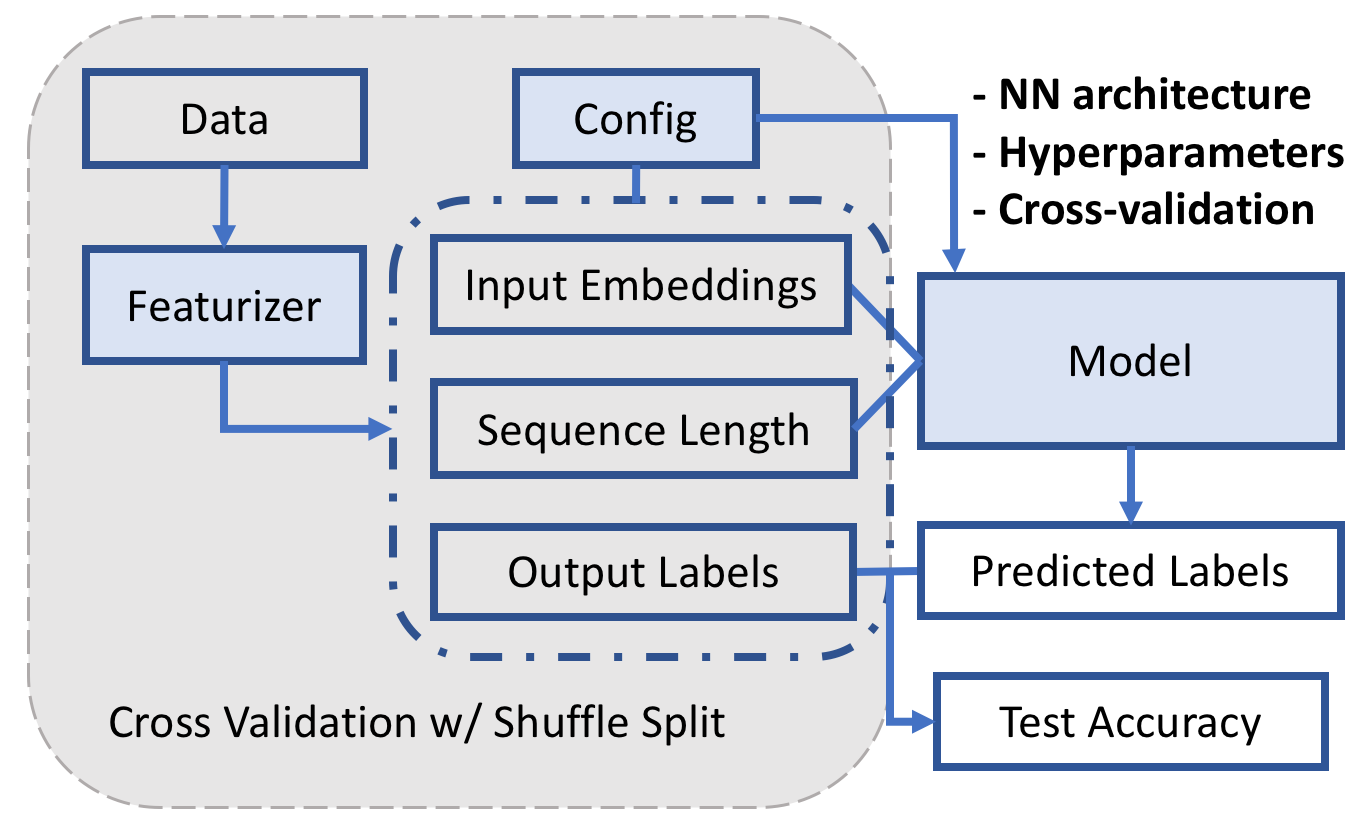}
    \caption{Hyperparameter search pipeline.  The data featurizer returns an object with attributes that can be saved and loaded into a model.  Cross-validation is performed via the config object, which specifies the network architecture, relevant hyperparameters, train/test/prediction datasets, etc.}
    \label{fig:pipeline}
\end{figure}

Exploring the efficacy of a variety of models and datasets requires an efficient, modular pipeline for standardizing data, specifying models, performing hyperparameter searches, and evaluating optimized models.  The first stage of this pipeline is data pre-processing. At this step, the data and sentiments are input in their raw forms and are output as lists of sentence vectors and class labels. To achieve this, we built a ``feature processor'' for the fully-labeled data which reads in a file (usually csv) and saves standardized input embeddings, sequence lengths, and output labels mapped to three sentiment classes (see Fig. \ref{fig:pipeline}).

Next, a training script reads in a text file specifying hyperparameters and creates a corresponding config object with those specifications.  A model object is initiated with a config object, and builds the appropriate NLP model.  The model object contains the requisite functions to build the Tensorflow graph, train the model, and perform predictions.  The training script also evaluates the model on a test set and saves the results (see Fig. \ref{fig:pipeline}). The models we implemented for this codebase include single-cell GRU, bi-directional GRU, multi-layer GRU, multi-layer bi-directional GRU, single-cell LSTM, bi-directional LSTM, multi-layer LSTM, multi-layer bi-directional LSTM, and (text) CNN.  We also have implemented the attention mechanism for each of the RNN models, but did not include it in our analysis due to time constraints.  

For our baseline, we classified sentiment using SVM (kernel and regularized), logistic regression (kernel and regularized), and dense neural networks (regularized).  The best cross-validated models across each hyperparameter search were saved.

\begin{center}
\begin{tabular}{ |c|c|c|c|c| }
    \hline
    Model/Dataset & Agricultural & GOP & Rotten Tomatoes & Self Driving Cars \\
    \hline
    SVM & 36\% & 61\% & 36\% & 61\% \\
    Logistic Regression & 33\% & 61\% & 33\% & 11\% \\
    Dense Neural Network & 53\% & 65\% & 53\% & 65\% \\
    \hline
\end{tabular}
\end{center}

Altogether, we wrote more than 2500 lines of code (Tensorflow and Python) from scratch and adapted more than 1500 lines of code from available resources.  Note that the CNN code in particular is heavily indebted to the work of \cite{britz2015}.

\section{Experiments}

\textit{Hyperparameter Search}---The config object makes it straightforward to perform hyperparameter searches.  We used GPU resources to run  14 distinct hyperparameter search campaigns (see supplementary materials).  Optimized hyperparameters included neural network model, hidden size (for RNNs), batch size, dropout probability, learning rate, L2 regularization lambda (for CNN), the number of filters (for CNN), and the number of training epochs.  Config parameters that we kept fixed were using dense rather than one-hot representations, always padding the inputs to a fixed length, using the same convolutional filter sizes (for CNN), using a fixed number of RNN layers (for multi-layer RNNs), and always clipping gradients with a maximum gradient norm of 5.  Nearly all of our models use 50-dimensional GloVe vectors.  In total, we have over 720 distinct hyperparameter runs (see the supplementary material).  To evaluate the performance of a given hyperparameter run, we perform three iterations of cross-validation with shuffle split for each hyperparameter set, and report the median accuracy and IQR.

\begin{figure}[h!]
    \centering
    \begin{subfigure}[b]{0.3\textwidth}
        \includegraphics[width=\textwidth]{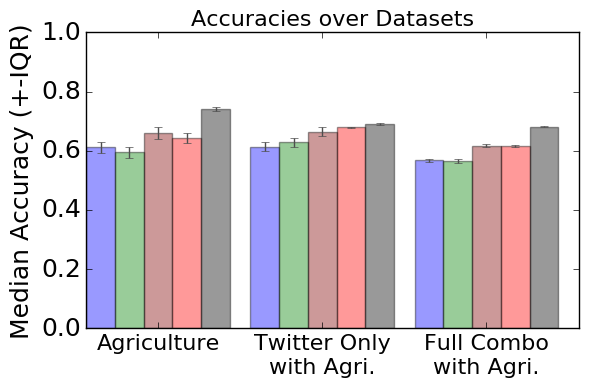}
        \caption{Accuracy comparison}
        \label{fig:hp_accuracy}
    \end{subfigure}
    ~ 
    \begin{subfigure}[b]{0.3\textwidth}
        \includegraphics[width=\textwidth]{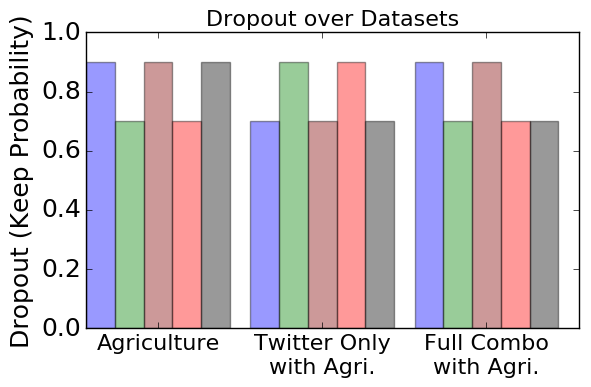}
        \caption{Dropout comparison}
        \label{fig:hp_dropout}
    \end{subfigure}
    ~ 
    \begin{subfigure}[b]{0.3\textwidth}
        \includegraphics[width=\textwidth]{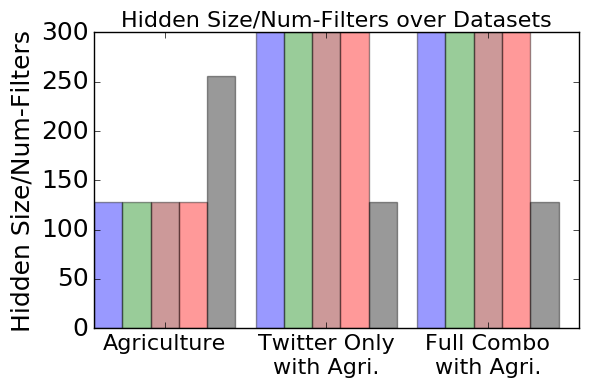}
        \caption{Hidden size comparison}
        \label{fig:hp_hidden}
    \end{subfigure}
    \caption{A sample comparison of model performance and correlated hyperparameters.  Blue is a single-cell GRU, green is a single-cell LSTM, brown is a bi-directional GRU, red is a bi-directional LSTM, and gray is a single-layer CNN.}\label{fig:hyperparams}
\end{figure}

\textit{Insights}---The main takeaway from the hyperparameter search was that CNNs performed better than the RNN models (see Fig.~ \ref{fig:hp_accuracy}).  It also turns out that on both CPUs and GPUs, CNN models train much faster (5 min) than RNN models (~1 hr).  This is not necessarily unexpected, as the expected advantages from RNN recurrence structure are smaller when sentences are smaller (as in tweets), and CNN filters are able to directly cover a larger proportion of smaller input strings.  Additionally, it became apparent that the dropout keep probability could be set anywhere between 0.5 and 0.9, depending on the model and number of hidden units or filters (see Fig. \ref{fig:hp_dropout}).  For RNNs, small datasets (Agriculture) require a smaller hidden size while the large augmented datasets (Twitter Only w/ Ag and Full Combo w/ Ag) require larger hidden size to achieve optimum performance (see Fig. \ref{fig:hp_hidden}).  This increase in hidden layer size suggests that the wider range of topics contained in the larger augmented datasets makes it more difficult to learn patterns generalizable to the target dataset.  


\begin{figure}
	\centering
	\includegraphics[width=0.6\textwidth]{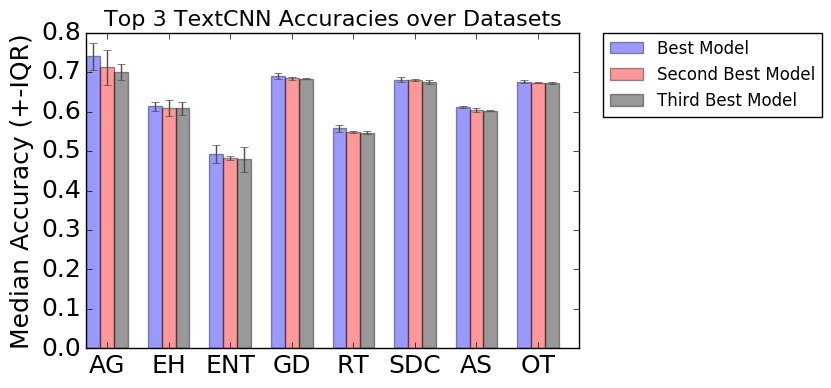}
    \caption{Accuracies of top three single-layer CNN models. AG, GD, RT, and SDC refer to previously described datasets. AS and OT were initial labels that correspond to to FC and TO, respectively. EH and ENT correspond to prelabeled sentiment datasets from CrowdFlower concerning economic health and economic news tone that were removed from further analysis due to low accuracies during hyperparameter searches (above) as well as small dataset size and sentence length 5-10x longer than the retained datasets.}
    \label{fig:cnn_acc}
\end{figure}

Given that single-layer CNNs broadly perform better than any of the RNN variants, we focused our attention on this model and isolated the top three models from the hyperparameter search for further analysis.  Overall, there was very little difference in accuracy among the top three models - the IQRs are extremely small, and all accuracies appear to fall nearly within each others' IQR ranges, which means our results are consistent (see Fig. \ref{fig:cnn_acc}).  Due to time and computation constraints, we therefore felt comfortable removing our three-fold cross validation and simply using a single model to evaluate CNN performance moving forward.

The optimized set of hyperparameters used in our final CNN models were as follows: (1) dropout keep probability = 0.5; (2) learning rate = 0.001; (3) number of filters per filter size = 64; (4) Filter sizes = [1, 2, 3, 4, 5, 10]; (5) Batch size = 100. Our dropout of 0.5 seems consistent with the results from Kim \cite{kim2014}, which show that CNNs have a substantial tendency to overfit the training data.  We also noticed a slight improvement (1-2\%) using 300-dimensional GloVe vectors compared with 50-dimensional GloVe vectors.

\begin{figure}[h!]
    \centering
    \begin{subfigure}[b]{0.3\textwidth}
        \includegraphics[width=\textwidth]{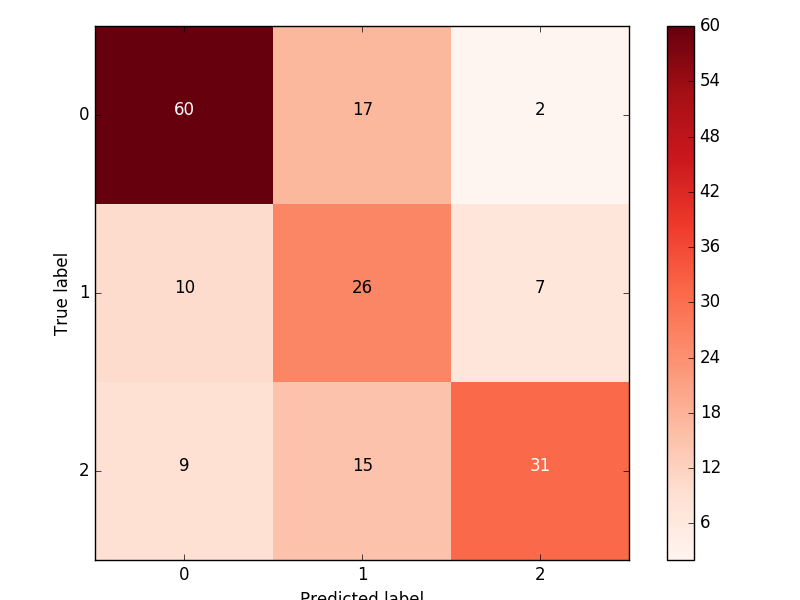}
        \caption{Confusion matrix on test set of Agriculture dataset, using the final trained model.}
        \label{fig:conf_mat}
    \end{subfigure}
    ~ 
    \begin{subfigure}[b]{0.33\textwidth}
        \includegraphics[width=\textwidth]{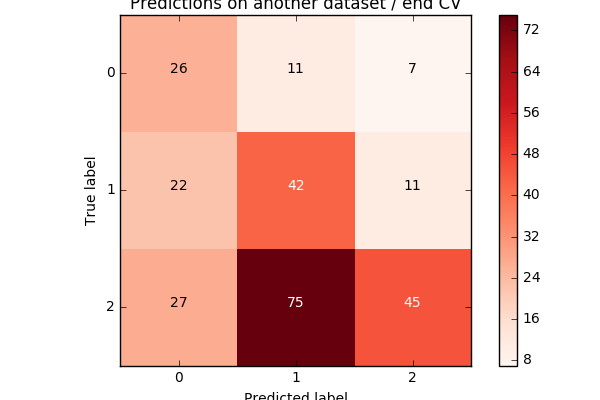}
        \caption{Confusion matrix on Kansas wheat, using the model yielding the best development accuracy.}
        \label{fig:conf_mat_ag_best_dev}
    \end{subfigure}
    \caption{Confusion matrices for Agriculture (AG) test set and Kansas wheat (KWT) dataset.}
    \label{fig:confusion_matrices}
\end{figure}

Our final CNN models were then used to predict sentiment on the held out Kansas wheat Twitter dataset.  The first confusion matrix in Fig. \ref{fig:confusion_matrices} describes the train set at the end of the training process.  The model is clearly able to distinguish between positive and negative sentiments, although it has a more difficult time with neutral ones.  Even more noticeable is the performance on the held-out Kansas wheat dataset, where the majority of positive tweets were predicted to be neutral.

\begin{figure}[h!]
	\centering
	\includegraphics[width=0.55\textwidth]{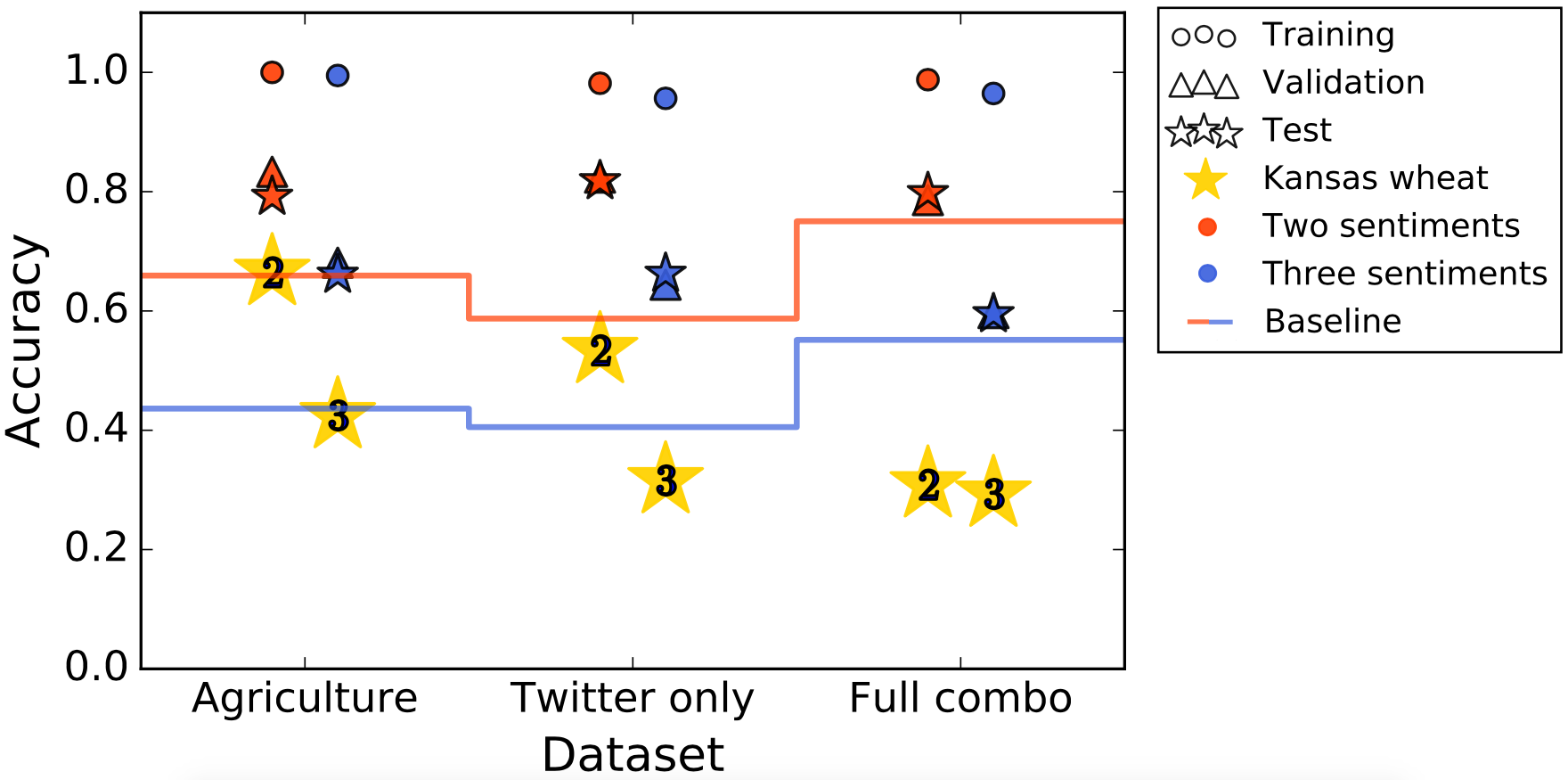}
    \caption{Model training, validation, and test performance and performance on Kansas Wheat Test (KWT) set for models trained on three different datasets for both 2- and 3-sentiment classes. The red and blue shapes represent the training, validation, and test accuracies for training a model with two and three sentiment classes, respectively. The gold stars represent predictions on KWT for two and three sentiments. The red and blue lines represent the baseline accuracies for non-deep learning models on KWT for two and three sentiment classes, respectively.}
    \label{fig:compare_all}
\end{figure}

Reviewing the literature, we find that predicting on fine-grained sentiments which include a neutral label is indeed a difficult task.  For instance, Socher et al.~achieve 85.4\% accuracy when predicting only positive and negative classes on an entire sentence, but achieve only 45.7\% when predicting on very negative, negative, neutral, positive, and very positive \cite{socher2013}.  While we are interested in including neutral sentiments in practical settings where neutral sentiments cannot be removed from a classification dataset in preprocessing, it is informative to see what the effect of removing the neutral examples would have on model performance.  Thus, as a final comparison, we ran training and testing on the Agriculture, Twitter Only, and Full Combo datasets, as well as prediction on the held out Kansas wheat Twitter dataset, for both binary (negative, positive) and ternary (negative, neutral, positive) sentiments, as shown in Fig.~\ref{fig:compare_all}.

In each case, see that we can overfit our train set (over 95\%), and do reasonably well on both the dev and test sets drawn from the same distribution as the train set.  However, we do less well on predicting sentiment on the held out Kansas wheat Twitter dataset, which is to be expected.  Interestingly, we confirm the occasionally drastic difference in performance between predicting binary and ternary sentiments, which can be as high as 20\% in terms of accuracy.

\textit{Examples of Misclassification}---Many of the model's mistakes were due to incorrectly labeling examples as neutral.  Below are some examples of such mistakes:

\begin{center}
\begin{tabular}{ |c|c|l| }
    \hline
    Label & Predicted & Tweet \\
    \hline
    Positive & Neutral & kansas is enjoying 'once in a lifetime' wheat harvest \\
    Positive & Neutral & wheat harvest big healthy crop in kansas \\
    Positive & Neutral & clay kansas got of rain this morning wheat looks good\\ 
    & & fall tillage was going well \\
    \hline
\end{tabular}
\end{center}

Additional mistakes were likely due to the CNN focusing on the wrong keywords.  This could potentially be remedied by implementing attention.

\begin{center}
\begin{tabular}{ |c|c|l| }
    \hline
    Label & Predicted & Tweet \\
    \hline
    Positive & Negative & induced dust bowl update kansas farmers harvest\\ 
    & & record winter wheat crop \\
    Negative & Positive & thomas kansas wheat struggling with little recent\\
    & & rain some light mist today \\
    Negative & Positive & thomas kansas wheat getting excessively dry probably\\ 
    & & done growing for the season \\
    \hline
\end{tabular}
\end{center}

There were also some confusing tweets that contained one sentiment for an unrelated topic and the opposing sentiment for wheat.  These are difficult to label even for a human.

\begin{center}
\begin{tabular}{ |c|c|l| }
    \hline
    Label & Predicted & Tweet \\
    \hline
    Neutral & Positive & phillips kansas cattle still gaining well in the cold \\
     & & with the lots dry wheat hasn't changed much this week \\
    Positive & Negative & clay kansas wheat still looks good warm temps today are starting\\ 
    & & to dry things not enough mud to make feeding cattle a challenge \\
    \hline
\end{tabular}
\end{center}

Finally, we noticed that there was a non-negligible amount of human error, demonstrating the importance of a rigorous labeling process.  In the following cases, for instance, we believe the model's prediction is more appropriate than the human label.

\begin{center}
\begin{tabular}{ |c|c|l| }
    \hline
    Label & Predicted & Tweet \\
    \hline
    Positive & Neutral & wheat harvest wrapping up in southern kansas \\
    Positive & Neutral & all aboard wheat harvest featured on ag am in kansas \\
    Positive & Negative & rains keep kansas wheat harvest at a slow pace \\
    \hline
\end{tabular}
\end{center}

Unfortunately, our time constraints compelled us to split up dataset labeling among four people, with different people labeling different subsets of tweets separated in time.  In the future, examples to be labeled should be shuffled randomly before distribution, with appropriate steps taken to normalize across bias in each labeler.

\textit{Validation Infusion}---We briefly investigated a procedure we refer to as ``validation infusion'', in which we train a model on a dataset from a non-agriculture domain but instead of using a partition of the same dataset for validation and testing, we substitute some percentage of the validation set with the agricultural sentiment (AG) dataset and use a test set composed of only AG. In our formulation, the entirety of AG is used. First, some percentage of AG is set aside as the test set. Second, an appropriate partition of the unrelated (training) dataset is allocated to the validation dataset such that the percentage of the validation dataset that comprises AG data is the percentage pre-specified by the modeler. The idea is that even though the model is trained on data unrelated to agriculture, it is evaluated (at least partially) on agricultural sentiment data. We found that this validation infusion procedure did not change the accuracies of KWT prediction when compared to the same dataset without validation infusion. 

\textit{Salient Characteristics of RNN models}---Our RNN models work best for our dataset with small learning rates (best performance at $1$ x $10^{-4}$). Improvement ($\approx 2$\%) in accuracy is obtained when the GRUs/LSTMs are bi-directional and a further gain in performance is obtained when the GRUs/LSTMs are deeper (multi-layered). LSTM-based models seem to be slightly better than GRU. For an aggregated dataset, a hidden size of 300 seems to work best.

\section{Conclusions and Future Work}
This project sought to investigate the potential of both direct learning and transfer learning to predict sentiment about crop yield from tweets from farming communities.  For unrelated domains, we used publicly available, fully-labeled datasets on sentiment related to movie reviews, the first 2016 GOP debate, and self driving cars.  For the target domain, we downloaded a sizable amount of crop-related Twitter posts and manually filtered and labeled them for positive, negative, and neutral sentiment.  To determine the best neural network for this application, we built a robust pipeline for extensive hyperparameter search across models, ultimately running over 720 distinct models.  Our analysis revealed that CNNs outperform and train faster than RNN variants across datasets. We also found that training on smaller, relevant datasets outperformed training on larger, unrelated datasets.  Therefore, our initial hypothesis that NLP models could effectively transfer learning across domains was incorrect. Furthermore, we demonstrated that predicting ternary sentiments is much more difficult than predicting binary sentiments, which is backed up by the literature.  However, we believe that accurately predicting on at least negative, neutral, and positive classes is highly relevant to many practical applications, including crop yield prediction, wherein forcing a two-sentiment structure would add substantial (and unwarranted) noise to the data.  Therefore, more effort should be invested into building effective models for fine-grained sentiment analysis, not just binary classification. Addressing these challenges in the future requires more labeled data in the domain of interest -- we hope that this project can effectively motivate the creation of agriculturally-relevant social media datasets.  Optimally performing these prediction tasks may also require more sophisticated models, such as incorporating attention into both the GRU/LSTM variants and more layers into a CNN. 

\section*{Acknowledgments}

We would like to thank Chris Manning for teaching CS 224N and being our project mentor; Patrick Baylis and Sabelle Smythe for helping us with our Twitter queries; and Marshall Burke, Stefano Ermon, and David Lobell for their feedback on this project as part of the Stanford Sustainability and AI Lab.

\section*{Contributions}

JD, SG, DH, and BH envisioned the project, developed and tested hypotheses, ran experiments on GPU resources, analyzed results, discussed conclusions, and wrote the report. JD adapted and maintained the CNN code. SG wrote and maintained the baseline code.  SG and DH adapted and maintained the RNN code. DH analyzed model predictions on individual examples. BH processed the Twitter query, wrote and maintained the feauturization code, and wrote and maintained model processing and visualization code. 

\bibliographystyle{aaai}
\bibliography{main}

\end{document}